\documentclass{article} 
\usepackage{iclr2020_conference,times}

\usepackage{amsmath,amsfonts,bm}

\def\eqref#1{equation~\ref{#1}}

\def\1{\bm{1}}

\DeclareMathAlphabet{\mathsfit}{\encodingdefault}{\sfdefault}{m}{sl}
\SetMathAlphabet{\mathsfit}{bold}{\encodingdefault}{\sfdefault}{bx}{n}

\DeclareMathOperator*{\argmin}{arg\,min}

\usepackage{hyperref}
\usepackage{url}
\usepackage[thinc]{esdiff}
\usepackage[colorinlistoftodos]{todonotes}
\usepackage{amsmath}
\usepackage{algorithm}
\usepackage[noend]{algpseudocode}
\usepackage{mathtools}
\usepackage{booktabs}
\usepackage{wrapfig}

\DeclarePairedDelimiterX{\infdivx}[2]{(}{)}{%
  #1\;\delimsize\|\;#2%
}

\makeatletter
\def\BState{\State\hskip-\ALG@thistlm}
\makeatother

\title{Better Fine-Tuning by Reducing Representational Collapse}

\iclrfinalcopy
\author{Armen Aghajanyan, Akshat Shrivastava, Anchit Gupta \& Naman Goyal \\ 
Facebook\\
\texttt{\{armenag,akshats,anchit,naman\}@fb.com} \\
\And
Luke Zettlemoyer \& Sonal Gupta \\
Facebook \\
\texttt{\{lsz, sonalgupta\}@fb.com} \\
}

\begin{document}

\maketitle
\begin{abstract}
    Although widely adopted, existing approaches for fine-tuning pre-trained language models have been shown to be unstable across hyper-parameter settings, motivating recent work on trust region methods. In this paper, we present a simplified and efficient method rooted in trust region theory that replaces previously used adversarial objectives with parametric noise (sampling from either a normal or uniform distribution), thereby discouraging representation change during fine-tuning when possible without hurting performance. We also introduce a new analysis to motivate the use of trust region methods more generally, by studying representational collapse; the degradation of generalizable representations from pre-trained models as they are fine-tuned for a specific end task. Extensive experiments show that our fine-tuning method matches or exceeds the performance of previous trust region methods on a range of understanding and generation tasks (including DailyMail/CNN, Gigaword, Reddit TIFU, and the GLUE benchmark), while also being much faster. We also show that it is less prone to representation collapse; the pre-trained models maintain more generalizable representations every time they are fine-tuned. 
\end{abstract}

\section{Introduction}

Pre-trained langauge models~\citep{GPT,BERT,ROBERTA, BART,MARGE} have been shown to capture a wide array of semantic, syntactic and world knowledge \citep{what_does_bert_look_at}, and provide the defacto initialization for modeling most existing NLP tasks. However, fine-tuning them for each task has been shown to be a highly unstable process, with many hyperparmeter settings producing failed fine-tuning runs, unstable results (large variation between random seeds), over-fitting and other unwanted consequences \citep{revisiting_bert, fine_tuning_data_order_dodge}. 

Recently, trust region or adversarial based based approaches, including SMART \citep{SMART} and FreeLB \citep{freelb}, have been shown to increase the stability and accuracy of fine-tuning, by adding extra constraints limiting how much the fine-tuning changes the initial parameters. However, these methods are significantly more computationally and memory intensive than the more commonly adopted simple-gradient-based approaches.

In this paper, we present a lightweight fine-tuning strategy which matches or improves performance relative to SMART and FreeLB, while needing just a fraction of the computational and memory overhead and no additional backwards passes. Our approach is motivated by trust region theory while also reducing to simply regularizing the model relative to parametric noise applied to the original pre-trained representations.  
We show uniformly better performance, setting a new  state of the art for RoBERTa fine-tuning on GLUE and reaching state of the art on XNLI using no novel pretraining approaches \citep{ROBERTA,GLUE,xnli}. Furthermore, the low overhead of our family of fine-tuning methods allows our method to be applied to generation tasks where we consistently outperform standard fine-tuning, setting state of the art on summarization tasks.

We also introduce a new analysis to motivate the use of trust-region-style methods more generally, by defining a new notion of representational collapse and introducing new methodology for measuring it during fine-tuning.
Representational collapse is \textbf{the degradation of generalizable representations of pre-trained models during the fine-tuning stage}. We empirically show that standard fine-tuning degrades generalizable representations through a series of probing experiments on GLUE tasks. Furthermore, we attribute this phenomena to using standard gradient descent algorithms for the fine-tuning stage. We also find that (1) recently proposed fine-tuning methods rooted in trust region, i.e. SMART, are capable of alleviating representation collapse, and (2) our methods alleviate representational collapse to an even great degree, manifesting in better performance across almost all datasets and models.

Our contributions in this paper are the following.
\begin{itemize}
    \item We propose a novel approach to fine-tuning rooted in trust-region theory which we show directly alleviates representational collapse at a fraction of the cost of other recently proposed fine-tuning methods.
    \item Through extensive experimentation, we show that our method outperforms standard fine-tuning methodology following recently proposed best practices from \cite{revisiting_bert}. We improve various SOTA models from sentence prediction to summarization, from monolingual to cross-lingual.
    \item We further define and explore the phenomena of representational collapse in fine-tuning and directly correlate it with generalization in tasks of interest.
\end{itemize}

\section{Learning Robust Representations through Regularized Fine-tuning}
We are interested in deriving methods for fine-tuning representations which provide guarantees on movement of representations, in the sense that they do not forget the original pre-trained representations when they are fine-tuned for a new tasks (see Section~\ref{rep_collapse} for more details). We introduce a new finetuning method rooted in an approximation to trust region, which provide guarantees for stochastic gradient descent algorithms by bounding some divergence between model at update $t$ and $t+1$ \citep{revisiting_natural_gradient, trust_region, SMART}.

Let $f: \mathbb{R}^{m \times n} \rightarrow \mathbb{R}^{p}$ be a function which returns some pre-trained representation parameterized by $\theta_f$ from $m$ tokens embedded into a fixed vector of size $n$.  Let the learned classification head $g: \mathbb{R}^{p} \rightarrow \mathbb{R}^{q}$ be a function which takes an input from $f$ and outputs a valid probability distribution parameterized by $\theta_g$ in $q$ dimensions. In the case of generation, we can assume the classification head is simply an identity function or softmax depending on the loss function.
Let $\mathcal{L}(\theta)$ denote a loss function given by $\theta=[\theta_f, \theta_g]$. We are interested in minimizing $\mathcal{L}$ with respect to $\theta$ such that each update step is constrained by movement in the representational density space $p(f)$. More formally
given an arbitrary $\epsilon$

\begin{align}
\begin{split}
\argmin_{\Delta \theta}&\ \mathcal{L}(\theta + \Delta\theta) \\
&\textit{s.t. } KL(p(f(\cdot \ ;\theta_f)) || p(f(\cdot \ ;\theta_f + \Delta\theta_{f}))) = \epsilon
\end{split}
\end{align}

This constrained optimization problem is equivalent to doing natural gradient descent directly over the representations \citep{revisiting_natural_gradient}. Unfortunately, we do not have direct access to the density of representations therefore it is not trivial to directly bound this quantity. Instead we propose to do natural gradient over $g \cdot f$ with an additional constraint that $g$ is at most 1-Lipschitz (which naturally constrains change of representations, see Section~\ref{section:theory} in the Appendix). Traditional computation of natural gradient is computationally prohibitive due to the need of inverting the Hessian. An alternative formulation of natural gradient can be stated through mirror descent, using Bregmann divergences \citep{mirror_descent_natural_gradient, SMART}. 

\begin{align}
    \mathcal{L}_{SMART}(\theta, f, g) = \mathcal{L}(\theta) + \lambda \mathbb{E}_{x \sim X}\left[\sup_{x^{\sim}: |x^{\sim} - x|\leq \epsilon}{KL_S\infdivx{g \cdot f(x)}{g \cdot f(x^{\sim})}}\right]
    \label{equation:BREG}
\end{align}

However, the supremum is computationally intractable. An approximation is possible by doing gradient ascent steps, similar to finding adversarial examples. This was first proposed by SMART with a symmetrical $KL_S(X,Y) = KL(X || Y) + KL(Y || X)$ term \citep{SMART}. 

We propose an even simpler approximation which does not require extra backward computations and empirically works as well as or better than SMART. We completely remove the adversarial nature from SMART and instead optimize for a smoothness parameterized by $KL_S$. Furthermore, we optionally also add a constraint on the smoothness of $g$ by making it at most 1-Lipschitz, the intuition being if we can bound the volume of change in $g$ we can more effectively bound $f$.

\begin{align}
    \mathcal{L}_{R3}(f,g,\theta) &= \mathcal{L}(\theta) + \lambda KL_S\infdivx{g \cdot f(x)}{g \cdot f(x + z)} && \text{\quad \quad \quad \ \ \ \textbf{R3F} Method}\\
    &s.t. \quad z \sim \mathcal{N}(0, \sigma^2 I) \text{ or } z \sim \mathcal{U}(-\sigma, \sigma) \label{eq:r3f_noise} \\
    &s.t. \quad Lip\{g\}\leq1 && \text{Optional \textbf{R4F} Method}
\end{align}
where $KL_S$ is the symmetric KL divergence and $z$ is a sample from a parametric distribution. In our work we test against two distributions, normal and uniform centered around $0$. We denote this as the \textbf{R}obust \textbf{R}epresentations through \textbf{R}egularized \textbf{F}inetuning (\textbf{R3F}) method.

Additionally we propose an extension to R3F (\textbf{R4F}; \textbf{R}obust \textbf{R}epresentations through \textbf{R}egularized and \textbf{R}eparameterized \textbf{F}inetuning, which reparameterizes $g$ to be at most 1-Lipschitz via Spectral Normalization \citep{spectral_normalization}. By constraining $g$ to be at most 1-Lipschitz, we can more directly bound the change in representation (Appendix Section~\ref{section:theory}). Specifically we scale all the weight matrices of $g$ by the inverse of their largest singular values $W_{SN} \coloneqq W/\sigma(W)$. Given that spectral radius $\sigma(W_{SN})=1$ we can bound $Lip\{g\} \leq 1$. In the case of generation, $g$ does not have any weights therefore we can only apply the R3F method.

\subsection{Relationship to SMART and FreeLB}
Our method is most closely related to the SMART algorithm which utilizes an auxiliary smoothness inducing regularization term which directly optimizes the Bregmann divergence mentioned above in Equation~\ref{equation:BREG} \citep{SMART}.

\begin{wraptable}{r}{5.5cm}
\centering
\begin{tabular}{@{}llll@{}}
\toprule
         & FP & BP & xFP \\ \midrule
FreeLB   & $1+S$              & $1+S$               & $3+3S$       \\
SMART    & $1+S$              & $1+S$               & $3+3S$       \\
R3F/R4F  & $2$              & 1               & 4        \\
Standard & $1$              & 1               & 3        \\ \bottomrule
\end{tabular}
\caption{Computational cost of recently proposed fine-tuning algorithms. We show Forward Passes (FP), Backward Passes (BP) as well as computation cost as a factor of forward passes (xFP). $S$ is the number of gradient ascent steps, with a minimum of $S\geq1$}
\label{table:computational_cost}
\end{wraptable}

SMART solves the supremum by using an adversarial methodology to ascent to the largest KL divergence with an $\epsilon-$ball. We instead propose to remove the ascent step completely, optionally fixing the smoothness of the classification head $g$. This completely removes the adversarial nature of SMART and is more akin to optimizing the smoothness of $g \cdot f$ directly.  Another recently proposed adversarial method for fine-tuning, FreeLB optimizes a direct adversarial loss $\mathcal{L}_{FreeLB}(\theta) = \sup_{\Delta\theta: |\Delta\theta|\leq\epsilon} \mathcal{L}(\theta + \Delta\theta)$ through iterative gradient ascent steps. Unfortunately the need for extra forward-backward passes can be prohibitively expensive when finetuning large pre-trained models \citep{freelb}.

Our method is significantly more computationally efficient than adversarial based fine-tuning methods, as seen in Table~\ref{table:computational_cost}. 
We show that this efficency does not hurt performance, 
we are able to match or exceed FreeLB and SMART on a large amount of tasks. In addition, the relatively low costs of our methods allows us to improve over fine-tuning on an array of generation tasks.

\section{Experiments}
We will first measure performance by fine-tuning on a range of tasks and languages. The next sections report analysis as to why methods rooted in trust region, including ours, outperform standard fine-tuning. Throughout all of our experiments, we aimed for fair comparisons, by using fixed budget hyper-parameters searches across all methods. Furthermore for computationally tractable tasks we report median/max numbers as well as show distributions across a large number of runs.
\subsection{Sentence Prediction}
\subsubsection*{GLUE}
\label{section:GLUE}
We will first test R3F and R4F on sentence classification tasks from the GLUE benchmark \citep{GLUE}. We select the same subset of GLUE tasks that have been reported by prior work in this space \citep{SMART}: MNLI \citep{mnli}, QQP \citep{qqp}, RTE \citep{rte}, QNLI \citep{qnli}, MRPC \citep{mrpc}, CoLA \citep{cola}, SST-2 \citep{sst2}.\footnote{We do not test against STS-B because it is a regression task where our KL divergence is not defined~\citep{stsb}.}

Consistent with prior work~\citep{SMART, freelb}, we focus on improving the performance of RoBERTa-Large based models in the single task setting \citep{ROBERTA}. We report performance of all models on the GLUE development set.

\begin{wrapfigure}{l}{0.5\textwidth} 
    \centering
    \includegraphics[width=0.5\textwidth]{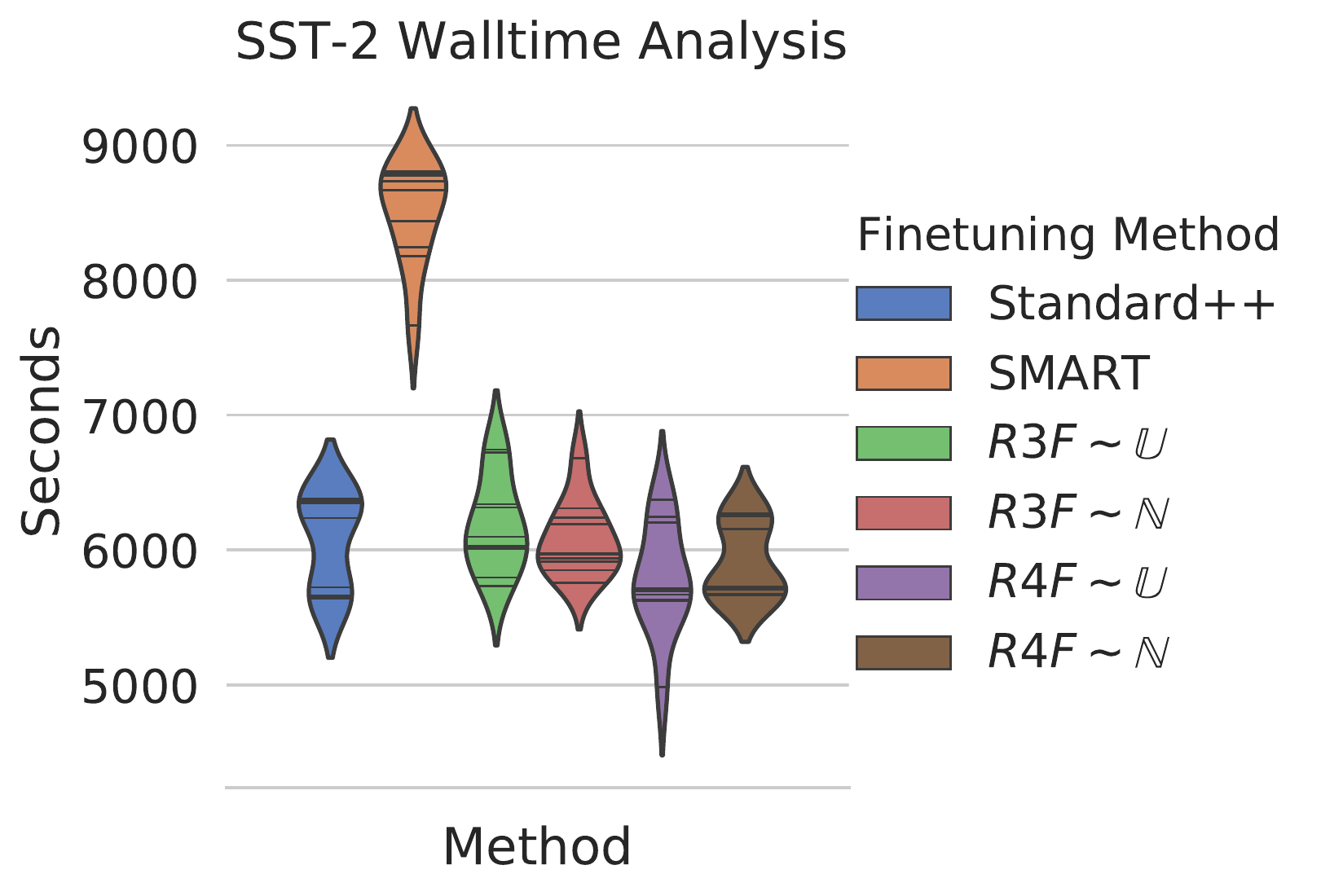}
    \caption{Empirical evidence towards the computational benefits of our method we present training wall time analysis on the SST-2 dataset. Each method includes a violin plot for 10 random runs. We define wall-time as the training time in seconds to best checkpoint.}
    \label{fig:sst2_walltime}
\end{wrapfigure}

We fine-tune each of the GLUE tasks with 4 methods: Standard (STD), the traditional fine-tuning scheme as done by RoBERTa \citep{ROBERTA}; Standard++ (STD++), a variant of standard fine-tuning that incorporates recently proposed best practices for fine-tuning, specifically longer fine-tuning and using bias correction in Adam \citep{revisiting_bert}; and our proposed methods R3F and R4F. We compare against the numbers reported by SMART, FreeLB and RoBERTa on the validation set. For each method we applied a hyper-parameter search with equivalent fixed budgets per method. Fine-tuning each task has task specific hyper-parameters described in the Appendix (Section~\ref{section:glue_hp}). After finding the best hyper-parameters we replicated experiments with optimal parameters across 10 different random seeds. Our numbers reported are the maximum of 10 seeds to be comparable with other benchmarks in Table~\ref{table:glue}. 

\begin{figure}
    \centering
    \includegraphics[width=\textwidth]{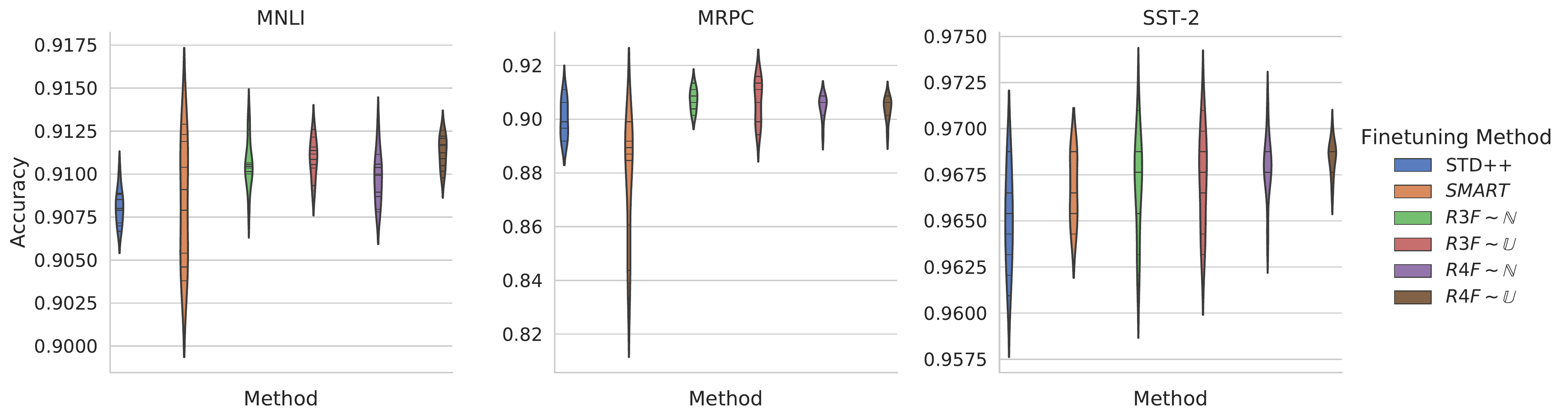}
    \caption{We show the results of our method against Standard++ fine-tuning and SMART across 3 tasks. Across 10 random seeds both \textit{max} and \textit{median} of our runs were higher using our method than both SMART and Standard++.}
    \label{fig:glue_stability}
\end{figure}

In addition to showing best performance, we also show the distribution of various methods across 10 seeds to demonstrate the stability properties of individual methods in Figure~\ref{fig:glue_stability}.

\begin{table}[h]
\scriptsize\setlength{\tabcolsep}{3pt}
\begin{tabular}{@{}llllllll@{}}
\toprule
 &
  \begin{tabular}[c]{@{}l@{}}MNLI\\ Acc-m/mm\end{tabular} &
  \begin{tabular}[c]{@{}l@{}}QQP\\ Acc/F1\end{tabular} &
  \begin{tabular}[c]{@{}l@{}}RTE\\ Acc\end{tabular} &
  \begin{tabular}[c]{@{}l@{}}QNLI\\ Acc\end{tabular} &
  \begin{tabular}[c]{@{}l@{}}MRPC\\ Acc\end{tabular} &
  \begin{tabular}[c]{@{}l@{}}CoLA\\ Mcc\end{tabular} &
  \begin{tabular}[c]{@{}l@{}}SST-2\\ Acc\end{tabular} \\ \midrule

STD   & 90.2/-    & 92.2/-    & 86.6 & 94.7 & 89.1    & 68.0 & 96.4    \\
STD++ & 91.0/-   & 92.2/-    & 87.4      &  94.8    & 91.1    &  69.4   &  96.9           \\
FreeLB      & 90.6/- & \textbf{92.6}/- & 88.1 & 95.0 & - & 71.1 & 96.7\\
SMART      & \textbf{91.1/91.3} & 92.4/89.8 & \textbf{92.0} & \textbf{95.6} & 89.2 & 70.6 & 96.9\\\midrule
R3F        & \textbf{91.1/91.3}         &  92.4/89.9         &  88.5    &     95.3  &   \textbf{91.6}        & \textbf{71.2}      &  97.0    \\
R4F        & 90.1/90.8  &   92.5/89.9        &  88.8    &   95.1   & 90.9    &  70.6    &   \textbf{97.1}   \\ \bottomrule

\end{tabular}
\quad
\begin{tabular}{@{}lllllll@{}}
\toprule
  \begin{tabular}[c]{@{}l@{}}MNLI\\ Acc-m/mm\end{tabular} &
  \begin{tabular}[c]{@{}l@{}}QQP\\ Acc/F1\end{tabular} &
  \begin{tabular}[c]{@{}l@{}}RTE\\ Acc\end{tabular} &
  \begin{tabular}[c]{@{}l@{}}QNLI\\ Acc\end{tabular} &
  \begin{tabular}[c]{@{}l@{}}MRPC\\ Acc\end{tabular} &
  \begin{tabular}[c]{@{}l@{}}CoLA\\ Mcc\end{tabular} &
  \begin{tabular}[c]{@{}l@{}}SST-2\\ Acc\end{tabular} \\ \midrule

90.2/-    & 91.9/-    & 86.6 & 92.1 & 84.4    & 66.2 & 96.4    \\
90.8/-   & 92.1/-    & 87.4      & 92.5     & 89.1    &  68.4   &  96.9           \\
-/- & -/- & - & - & - & - & -\\
90.85/\textbf{91.10} & 91.7/88.2 & \textbf{89.5} & 94.8 & 83.9 & 69.4 & 96.6\\\midrule
\textbf{91.10/91.10}         &   \textbf{92.1/88.4}       &  88.4    & \textbf{95.1}  &   \textbf{91.2}        & \textbf{70.6}      &  96.2    \\
90.0/90.6  &    91.8/88.2       &  88.3    &  94.8    & 90.1    &  70.1    &   \textbf{96.8}   \\ \bottomrule

\end{tabular}
\caption{We present our best results on the GLUE development set for various fine-tuning methods applied to the RoBERTa Large model. On the left side table we present our best numbers and numbers published in other papers. On the right side we present median numbers from 10 runs for mentioned methods.}
\label{table:glue}
\end{table}

R3F and R4F unanimously improve over Standard and Standard++ fine-tuning. Furthermore our methods match or exceed adversarial methods such as SMART/FreeLB at a fraction of the computational cost when comparing median runs. We show computational cost in Figure~\ref{fig:sst2_walltime} for a single task, but the relative behavior of wall times are consistent across all other tasks in GLUE.
\subsubsection*{XNLI}
We hypothesize that staying closer to the original representations is especially important for cross-lingual tasks, especially in the zero-shot fashion where drifting away from pre-trained representations for a single language might manifest in loss of cross-lingual capabilities. In particular we take a look at the popular XNLI benchmark, containing 15 languages \citep{xnli}. We compare our method against the standard trained XLM-R model in the zero-shot setting \citep{xlmr}.


\begin{table}[h]
\centering\small
\setlength{\tabcolsep}{3pt}
    \begin{tabular}{@{}lllllllllllllllll@{}}
    \toprule
    Model             & en   & fr   & es   & de   & el   & bg   & ru   & tr   & ar   & vi   & th   & zh   & hi   & sw   & ur   & Avg  \\ \midrule
    XLM-R Base        & 85.8 & 79.7 & 80.7 & 78.7 & 77.5 & 79.6 & 78.1 & 74.2 & 73.8 & 76.5 & 74.6 & 76.7 & 72.4 & 66.5 & 68.3 & 76.2 \\
    XLM-R Large       & 89.1 & 84.1 & 85.1 & 83.9 & 82.9 & 84.0 & 81.2 & 79.6 & 79.8 & 80.8 & 78.1 & 80.2 & 76.9 & \textbf{73.9} & 73.8 & 80.9 \\
    \quad + R3F & 89.4 & 84.2 & 85.1 & 83.7 & 83.6 & 84.6 & 82.3 & \textbf{80.7} & \textbf{80.6} & \textbf{81.1} & \textbf{79.4} & 80.1 & 77.3 & 72.6 & 74.2 & 81.2 \\
    \quad + R4F & \textbf{89.6} & \textbf{84.7} & \textbf{85.2} & \textbf{84.2} & \textbf{83.6} & \textbf{84.6} & \textbf{82.5} & 80.3 & 80.5 & 80.9 & 79.2 & \textbf{80.6} & \textbf{78.2} & 72.7 & \textbf{73.9} & \textbf{81.4} \\\midrule
    InfoXLM & 89.7 & 84.5 & 85.5 & 84.1 & 83.4 & 84.2 & 81.3 & 80.9 & 80.4 & 80.8 & 78.9 & 80.9 & 77.9 & 74.8 & 73.7 & 81.4\\\bottomrule
    \end{tabular}
\caption{Average of 5 runs of zero-shots results on the XNLI test set for our method applied to XLM-R Large. Variouns of our method win over the majority of languages. The bottom row shows the current SOTA on XNLI which requires the pre-training of novel model.}
\label{table:xnli}
\end{table}

We present our result in Table~\ref{table:xnli}. R3F and R4F dominate standard pre-training on 14 out of the 15 languages in the XNLI task. R4F improves over the best known XLM-R XNLI results reaching SOTA with an average language score of 81.4 across 5 runs. The current state of the art required a novel pre-training method to reach the same numbers as \citep{infoxlm}.

\subsection{Summarization}
While prior work in non-standard finetuning methods tends to focus on sentence prediction and GLUE tasks \citep{SMART, freelb, revisiting_bert}, we look to improve abstractive summarization, due to its additional complexity and computational cost, specifically we look at three datasets: \textit{CNN/Dailymail} \citep{cnndailymail}, \textit{Gigaword} \citep{gigaword} and \textit{Reddit TIFU} \citep{reddittifu}. 

Like most other NLP tasks, summarization recently has been dominated by fine-tuning of large pre-trained models. For example PEGASUS explicitly defines a pre-training objective to facilitate the learning of representations tailored to summarization tasks manifesting in state-of the art performance on various summarization benchmarks \citep{pegasus}. ProphetNet improved over these numbers by introducing their own novel self-supervised task \citep{prophetnet}.

Independently of the pre-training task, standard fine-tuning on downstream tasks follows a simple formula of using a label smoothing loss while directly fine-tuning the whole model, without addition of any new parameters. We propose the addition of the R3F term directly to the label smoothing loss.

\begin{table}[]
\centering
\begin{tabular}{@{}llll@{}} \toprule
                   & CNN/DailyMail     & Gigaword          & Reddit TIFU (Long)      \\ \midrule
Random Transformer & 38.27/15.03/35.48 & 35.70/16.75/32.83 & 15.89/1.94/12.22 \\
BART               & 44.16/21.28/40.90 & 39.29/20.09/35.65 & 24.19/8.12/21.31  \\
PEGASUS            & 44.17/\textbf{21.47}/41.11 & 39.12/19.86/36.24 & 26.63/9.01/21.60
\\ \midrule
ProphetNet (Old SOTA)  & 44.20/21.17/\textbf{41.30} & 39.51/20.42/\textbf{36.69} & -
\\ \midrule
BART+R3F (New SOTA) & \textbf{44.38/21.53/41.17} & \textbf{40.45/20.69/36.56} & \textbf{30.31/10.98/24.74}\\ \bottomrule
\end{tabular}
\caption{Our results on various summarization data-sets. We report Rouge-1, Rouge-2 and Rouge-L per element in table. Following PEGASUS, we bold the best number and numbers within 0.15 of the best.}
\label{table:summarization}
\end{table}

We present our results in Table~\ref{table:summarization}. Our method (R3F) outperforms standard fine-tuning across the board for three tasks across all of the variants of the ROUGE metric. Notably we improve \textit{Gigaword} and \textit{Reddit TIFU} ROUGE-1 scores by a point and 4 points respectively.

\section{Representational Collapse}
\label{rep_collapse}
Catastrophic forgetting, originally proposed as catastrophic interference, is a phenomena that occurs during sequential training where new updates interfere catastrophically with previous updates manifesting in forgetting of certain examples with respect to a fixed task \citep{catastrophic_interference}. Inspired by this work, we explore the related problem of representational collapse; \textbf{the degradation of generalizable representations of pre-trained models during the fine-tuning stage.} This definition is independent of a specific fine-tuning task, but is rather over the internal representations generalizabality over a large union of tasks. Another view of this phenomena is that fine-tuning collapses the wide range of information available in the representations into a smaller set needed only for the immediate task and particular training set.

Measuring such degradations is non-trivial. Simple metrics such as the distance between pre-trained representations and fine-tuned representations is not sufficient (e.g. adding a constant to the pre-trained representations will not change representation power, but will change distances). One approach would be to estimate mutual information of representations across tasks before and after fine-tuning, but estimation of mutual information is notoriously hard, especially in high-dimensions \citep{mut_info_max}.
We instead propose a series of probing experiments meant to provide us with empirical evidence of the existence of representation collapse on the GLUE benchmark \citep{GLUE}.

\subsection{Probing Experiments}

\subsubsection*{Probing Generalization of fine-tuned Representations}
To measure the generalization properties of various fine-tuning methodologies, we follow probing methodology by first freezing the representations from the model trained on one task and then fine-tuning a linear layer on top of the model for another task. By doing this form of probing we can directly measure the quality of representations learned by various fine-tuning methods, as well as how much they collapse when fine-tuned on a sequence of tasks.

\begin{figure}
    \centering
    \includegraphics[width=\textwidth]{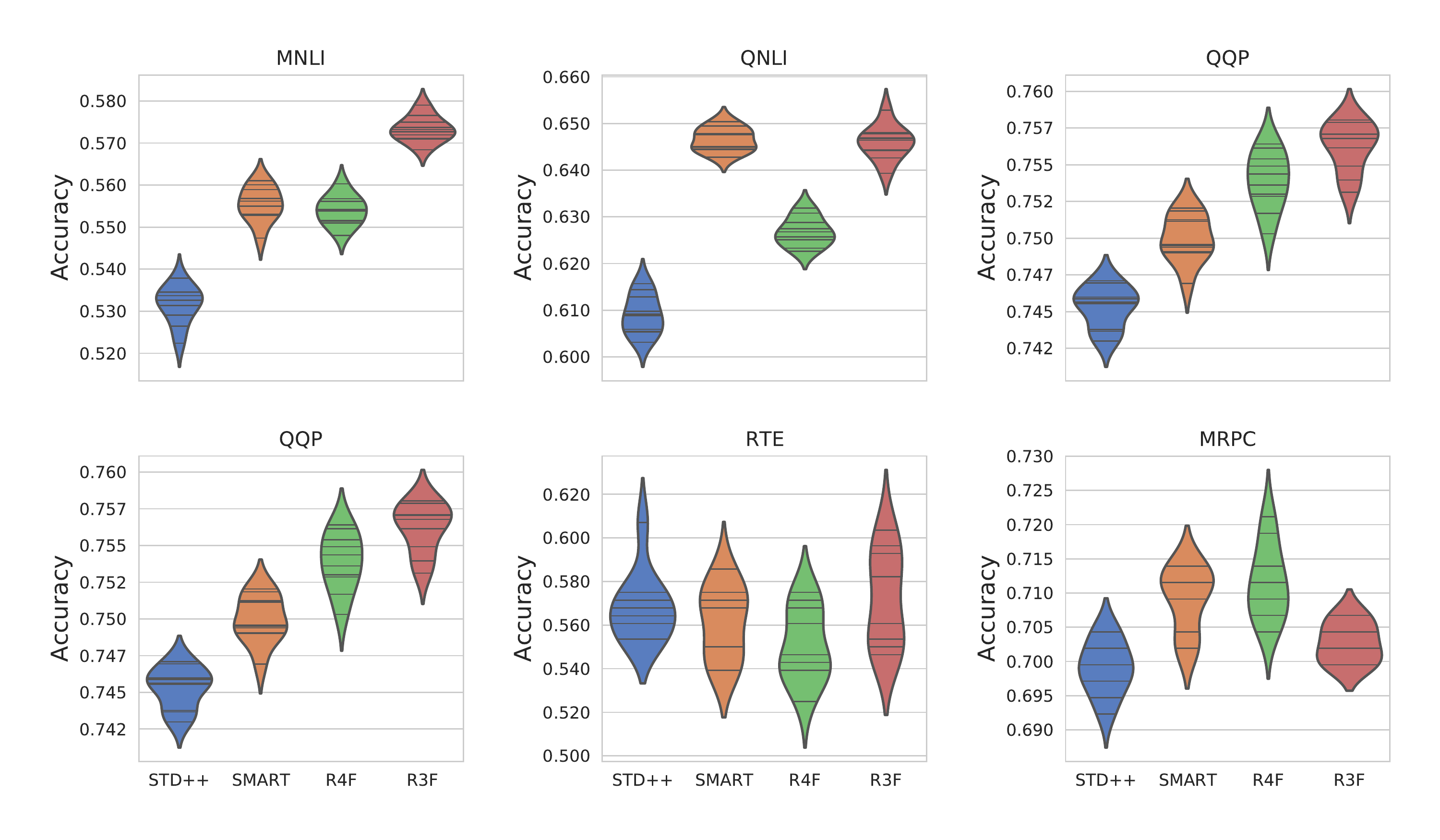}
    \caption{Results from our probing experiments comparing our proposed algorithms R3F, R4F to standard fine-tuning. Variants of our method consistently outperform past work.}
    \label{fig:probing}
\end{figure}

In particular, we finetune a RoBERTa model on SST-2 and train a linear layer for 6 other GLUE tasks respectively. Our results are shown in Figure~\ref{fig:probing}.  Appendix~\ref{section:glue_hp} presents the hyperparameters. Across all tasks, one of the two variants of our method performed best across various fine-tuning methods. Conversely standard fine-tuning produced representations which were worse than other fine-tuning methods across the board, hinting at the sub-optimality of standard fine-tuning. Furthermore R3F/R4F consistently outperforms the adversarial fine-tuning method SMART.

\subsubsection*{Probing Representation Degradation}

In order to show the effect of representation collapse, we propose an experiment to measure how the fine-tuning process degrades representations by sequentially training on a series of GLUE tasks. We arbitrarily select 3 GLUE tasks (QNLI, QQP, and RTE) and a source task (SST-2). We begin by training a model on our source task, and then train on QNLI, QQP, and RTE \begin{wrapfigure}{r}{0.5\textwidth} 
    \centering
    \vspace{-1.5em}
    \includegraphics[width=0.5\textwidth]{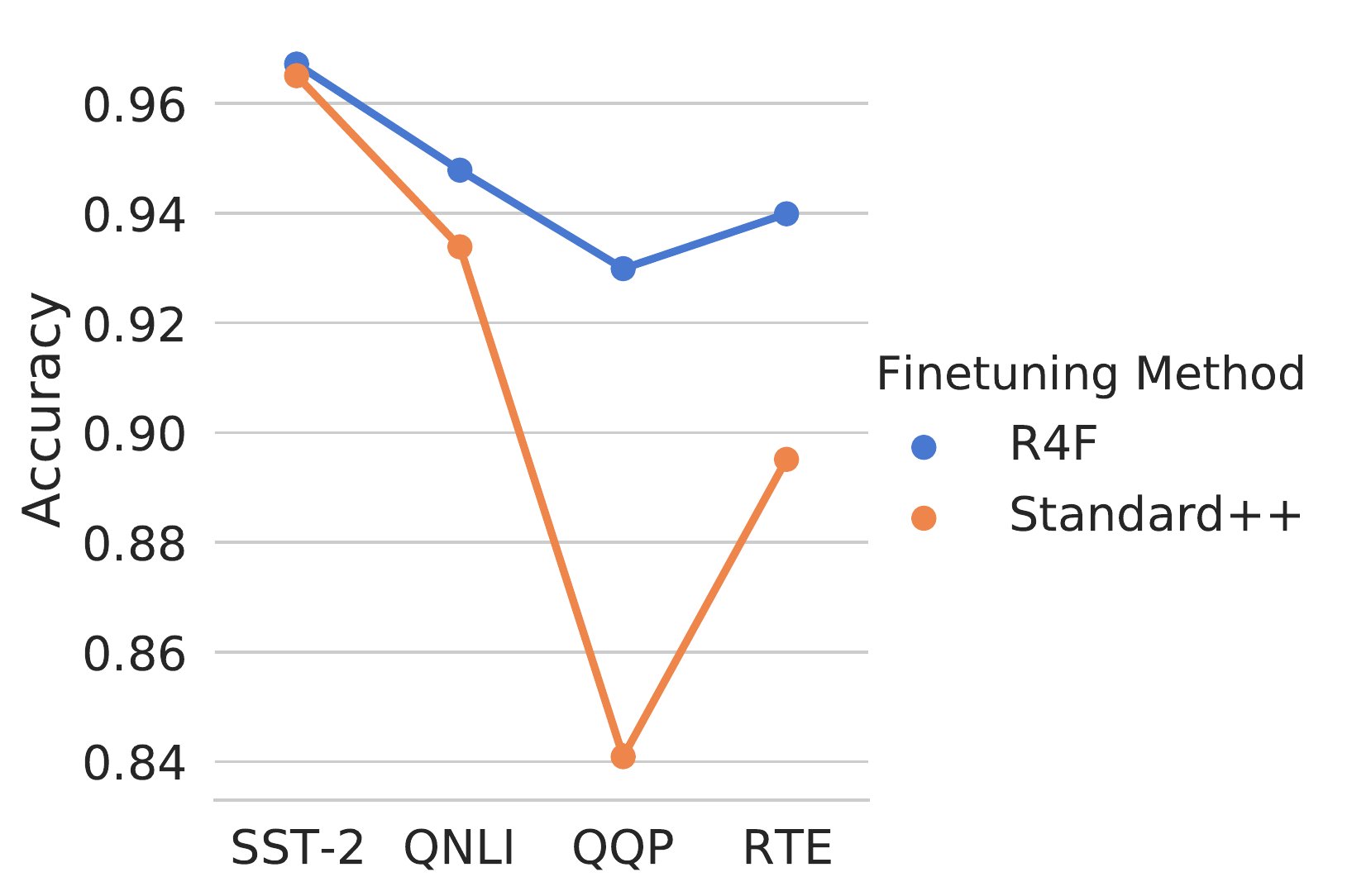}
    \caption{We show the results of the chained probing experiments. We do not show the distributional properties of the runs because there was very little variance in the results.}
    \label{fig:chain_probe}
    \vspace{-3.5em}
\end{wrapfigure}
in a sequential order using the best checkpoint from the prior iteration. At each point in the chain we probe the source task and measure performance. Our results are depicted in Figure~\ref{fig:chain_probe}.

As we can see with the standard fine-tuning process our model diverges from the source task resulting in lower performance probes, however with our method the probes vary much less with sequential probing resulting in better probing and end performance.

\subsubsection*{Probing Representation Retention}
To further understand the impact of representational collapse, we extend our probing experiments to train a cyclic chain of tasks. In our prior experiments we showed that traditional fine-tuning degrades representations during the fine-tuning process, meaning standard fine-tuning learns poorer representation compared to alternative fine-tuning methods. The dual to looking at degradation is to look at the retainment of learned representations, to do this we take a look at cyclic sequential probing. Sequential probing involves training a model on task A, probing B, then training model fine-tuned on B and probing task C, and so forth. We then create a cyclic chain $\underbrace{A \rightarrow B \rightarrow C}_\text{Cycle 1} \rightarrow \underbrace{ A \rightarrow B \rightarrow C}_\text{Cycle 2}$ from where we compare tasks via their probe performance at each cycle. 

We expect probing performance to increase at every cycle, since every cycle the task we are probing on will undergo a full fine-tuning. What we are interested in is the level of retention in representations after the fine-tuning. Specifically we hypothesize that our method, specifically R4F will retain representations significantly better than the Standard++ fine-tuning method. 

In our experiments we consider the following sequence of GLUE tasks:  SST-2 $\rightarrow$ QNLI $\rightarrow$ QQP $\rightarrow$ RTE. We defer hyperparameter values to Appendix (Section~\ref{section:glue_hp}). 

\begin{figure}[h]
    \centering
    \includegraphics[width=\textwidth]{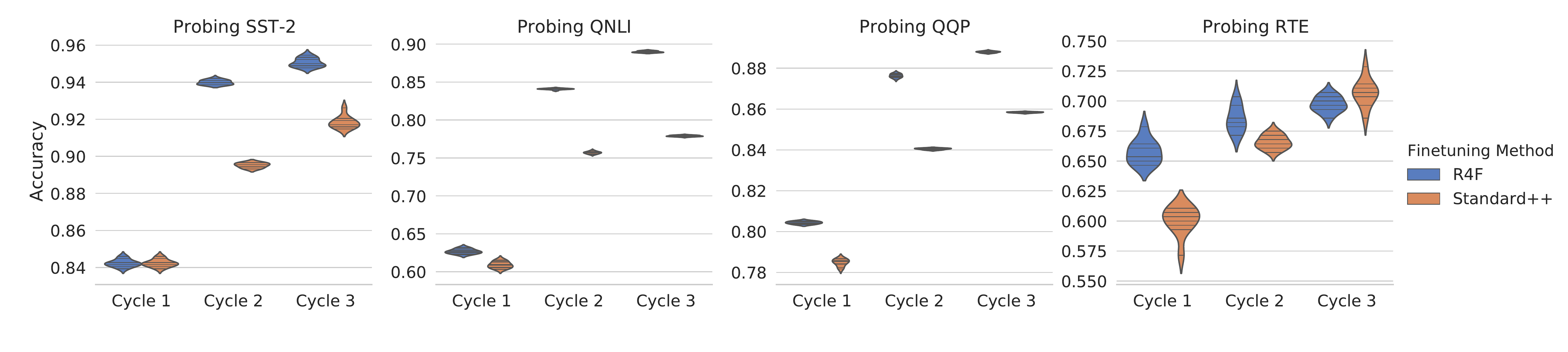}
    \caption{We present the results of cyclical sequential probing for 3 cycles.}
    \label{fig:glue_cyclic_probing}
\end{figure}

Looking at Figure~\ref{fig:glue_cyclic_probing}, we see that R4F retains quality of representations significantly better that standard fine-tuning methods.
\section{Conclusion}

We propose a family of new fine-tuning approaches for pre-trained representations based on trust-region theory: R3F and R4F. Our methods are more computationally efficient and out perform prior work in fine-tuning via adversarial learning \citep{SMART, freelb}. We show that this is due to a new phenomena that occurs during fine-tuning: representational collapse, where representations learned during fine-tuning degrade leading to worse generalization. Our analysis show standard fine-tuning is sub-optimal when it comes to learning generalizable representations, and instead our methods retain representation generalizability and improve end task performance. 

With our method we improve upon monolingual and multilingual sentence prediction tasks as well as generation tasks compared to standard and adversarial fine-tuning methods. Notably we set state of the art on DailyMail/CNN, Gigaword, Reddit TIFU, improve the best known results on fine-tuning RoBERTa on GLUE, and reach state of the art on zero-shot XNLI without the need for any new pre-training method.

\bibliography{iclr2020_conference}
\bibliographystyle{iclr2020_conference}

\appendix
\section{Appendix}

\subsection{Controlling Change of Representation via Change of Variable}
\label{section:theory}
Let us say we have random variables in some type of markovian chain $x,y,z; y = f(x ; \theta_f), z = g(y ; \theta_g)$

The change of variable formulation for probability densities is
\begin{equation}
    p(f(x ; \theta_f)) = p(g(f(x ; \theta_f)))\left|\det\frac{dg(f(x ; \theta_f))}{df(x ; \theta_f)}\right|  
\end{equation}

Direct application of change of variable gives us
\begin{align}
    KL(p(f(x ; \theta_f)) &|| p(f(x ; \theta_f + \Delta\theta_f))) = \\
    &\sum p(f(x ; \theta_f)) \log \frac{p(f(x ; \theta_f))}{p(f(x ; \theta_f + \Delta\theta_f))} =\\
    \sum p(g(f(x ; \theta_f)))&\left|\det\frac{dg(f(x ; \theta_f))}{df(x ; \theta_f)}\right| 
[\\
&\log p(g(f(x ; \theta_f))) + \log\left|\det\frac{dg(f(x ; \theta_f))}{df(x ; \theta_f)}\right| \\
-&\log p(g(f(x ; \Delta\theta_f))) - \log\left|\det\frac{dg(f(x ; \Delta\theta_f))}{df(x ; \Delta\theta_f)}\right| \\
&]
\end{align}

Let us make some more assumptions. Let $g(y) = Wy$ where the spectral norm of $W, \rho(W) = 1$. We can then trivially bound $\det W \leq 1$. Then we have

\begin{align}
 &= \sum p(g(f(x ; \theta_f)))\left|\det\frac{dg(f(x ; \theta_f))}{df(x ; \theta_f)}\right| [\log p(g(f(x ; \theta_f))) - \log p(g(f(x ; \Delta\theta_f))) ] \\
 &= \sum p(g(f(x ; \theta_f)))\left|\det\frac{dg(f(x ; \theta_f))}{df(x ; \theta_f)}\right| \log \frac{p(g(f(x ; \theta_f)))}{p(g(f(x ; \Delta\theta_f)))} \\
 &\leq \sum p(g(f(x ; \theta_f))) \log \frac{p(g(f(x ; \theta_f)))}{p(g(f(x ; \Delta\theta_f)))} \\
 &= KL(p(g(f(x ; \theta_f))) || p(g(f(x ; \Delta\theta_f))))
\end{align}

We also see that tightness is controlled by $|\det W|$ which is bounded by the singular value giving us intuition to the importance of using spectral normalization.  

\subsection{Experiment Hyper-Parameters}
\label{section:glue_hp}
For our GLUE related experiments both full fine-tuning and probing, the following parameters are used. For probing experiments the difference is our RoBERTa encoder is frozen and encoder dropout is removed.

\begin{table}[h]
\centering
\begin{tabular}{@{}llllllll@{}}
\toprule
Hyper Parameter & MNLI   & QNLI  & QQP    & SST-2 & RTE  & MRPC & CoLA \\ \midrule
Learning Rate   & 5e-6   & 5e-6  & 5e-6   & 5e-6  & 1e-5 & 1e-5 & 1e-5 \\
Max Updates     & 123873 & 33112 & 113272 & 20935 & 3120 & 2296 & 5336 \\
Max Sentences   & 8      & 8     & 32     & 32    & 8    & 16   & 16   \\
\bottomrule  
\end{tabular}
\caption{Task specific hyper parameters for GLUE experiments}
\end{table}

\begin{table}[h]
\centering
\begin{tabular}{@{}ll@{}}
\toprule
Hyper parameter & Value              \\ \midrule
Optimizer       & Adam               \\
Adam-betas      & (0.9, 0.98)        \\
Adam-eps        & 1e-6               \\
LR Scheduler    & polynomial decay   \\
Dropout         & 0.1                \\
Weight Decay    & 0.01               \\
Warmup Updates  & 0.06 * max updates \\
\bottomrule
\end{tabular}
\quad
\begin{tabular}{@{}ll@{}}
\toprule
Hyper parameter & Value           \\ \midrule
$\lambda$          & [0.1, 0.5, 1.0, 5.0] \\
Noise Types     & [$\mathcal{U}$, $\mathcal{N}$] \\
$\sigma$        & $1e-5$\\ 
\bottomrule
\end{tabular}
\caption{Hyper parameters for R3F and R4F experiments on GLUE}
\end{table}

\begin{table}[h]
\centering
\begin{tabular}{@{}llll@{}}
\toprule
Hyper Parameter & CNN/Dailymail   & Gigaword  & Reddit TIFU \\ \midrule
Max Tokens      & 1024 & 2048 & 2048 \\
Total updates   & 80000 & 200000 & 200000 \\
Warmup Updates  & 1000 & 5000 & 5000 \\
\bottomrule  
\end{tabular}
\caption{Task specific hyper parameters for Summarization experiments.}
\end{table}

\begin{table}[h]
\centering
\begin{tabular}{@{}ll@{}}
\toprule
Hyper parameter & Value              \\ \midrule
Optimizer       & Adam               \\
Adam-betas      & (0.9, 0.98)        \\
Adam-eps        & 1e-8               \\
LR Scheduler    & polynomial decay   \\
Learning Rate   & 3e-05              \\

\bottomrule
\end{tabular}
\quad
\begin{tabular}{@{}ll@{}}
\toprule
Hyper parameter & Value           \\ \midrule
$\lambda$          & [0.001, 0.01, 0.1] \\
Noise Types     & [$\mathcal{U}$, $\mathcal{N}$] \\
$\sigma$        & $1e-5$\\ 
Dropout         & 0.1                \\
Weight Decay    & 0.01               \\
Clip Norm       & 0.1                \\
\bottomrule
\end{tabular}
\caption{Hyper parameters for R3F and R4F experiments on Summarization experiments.}
\end{table}

\begin{table}[h]
\centering
\begin{tabular}{@{}ll@{}}
\toprule
Hyper parameter & Value              \\ \midrule
Optimizer       & Adam               \\
Adam-betas      & (0.9, 0.98)        \\
Adam-eps        & 1e-8               \\
LR Scheduler    & polynomial decay   \\
Learning Rate   & 3e-05              \\
Dropout         & 0.1                \\
Weight Decay    & 0.01               \\

\bottomrule
\end{tabular}
\quad
\begin{tabular}{@{}ll@{}}
\toprule
Hyper parameter & Value           \\ \midrule
$\lambda$          & [0.5, 1, 3, 5] \\
Noise Types     & [$\mathcal{U}$, $\mathcal{N}$] \\
$\sigma$        & $1e-5$\\ 
Total  Updates & 450000 \\
Max Positions   & 512 \\
Max Tokens      & 4400               \\
Max Sentences   & 8                  \\
\bottomrule
\end{tabular}
\caption{Hyper parameters for R3F and R4F experiments on XNLI.}
\end{table}

\end{document}